# GP-CNAS: Convolutional Neural Network Architecture Search with Genetic Programming

**Yiheng Zhu**

Engineering Mathematics
University of Bristol

**Yichen Yao, Zili Wu, Yujie Chen,
Guozheng Li, Haoyuan Hu & Yinghui Xu**

Artificial Intelligence Department
Zhejiang Cainiao Supply Chain Management Co., Ltd.

## Abstract

Convolutional neural networks (CNNs) are effective at solving difficult problems like visual recognition, speech recognition and natural language processing. However, performance gain comes at the cost of laborious trial-and-error in designing deeper CNN architectures. In this paper, a genetic programming (GP) framework for convolutional neural network architecture search, abbreviated as GP-CNAS, is proposed to automatically search for optimal CNN architectures. GP-CNAS encodes CNNs as trees where leaf nodes (GP terminals) are selected residual blocks and non-leaf nodes (GP functions) specify the block assembling procedure. Our tree-based representation enables easy design and flexible implementation of genetic operators. Specifically, we design a dynamic crossover operator that strikes a balance between exploration and exploitation, which emphasizes CNN complexity at early stage and CNN diversity at later stage. Therefore, the desired CNN architecture with balanced depth and width can be found within limited trials. Moreover, our GP-CNAS framework is highly compatible with other manually-designed and NAS-generated block types as well. Experimental results on the CIFAR-10 dataset show that GP-CNAS is competitive among the state-of-the-art automatic and semi-automatic NAS algorithms.

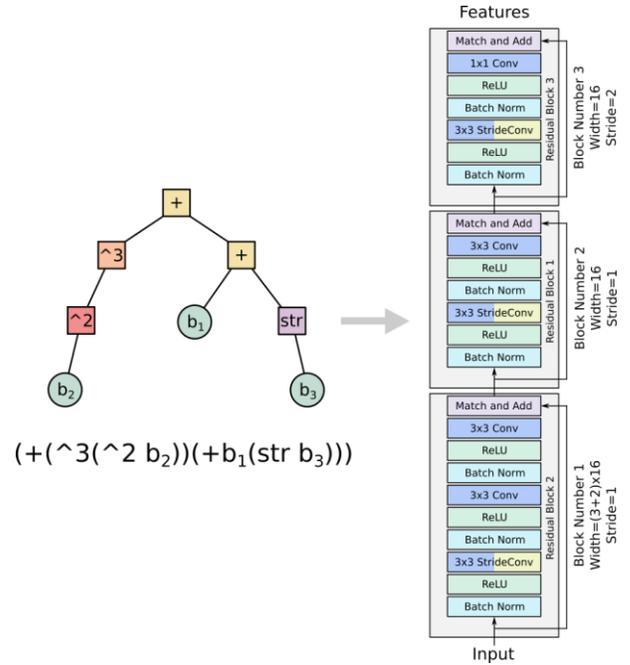

Figure 1: GP-CNAS encoding

## Introduction

Inspired by neuroscience studies (Hubel and Wiesel 1962), the convolutional neural network (CNN) was introduced by (LeCun et al. 1989) and popularized by their triumph in the ImageNet Large-Scale Visual Recognition Challenge (ILSVRC) (Russakovsky et al. 2015). Successful applications of CNNs to real-world problems include vision tasks, speech tasks, natural language processing and more (Chen et al. 2018; Xiong et al. 2017; Conneau et al. 2016).

In recent years, the design of CNN architectures favors increasing depth and width. The most noticeable work in the ILSVRC manifest the power of deeper CNNs: 8-layer AlexNet (Krizhevsky, Sutskever, and Hinton 2012), 19-layer VGGNet (Simonyan and Zisserman 2014) and 22-layer Inception (Szegedy et al. 2015). Recently, Srivastava, Greff, and Schmidhuber (2015) and He et al. (2016a) introduced the shortcut connection to the block-based CNN architecture, which enables the construction of even deeper architectures like the 152-layer residual network (ResNet). Goodfellow et al. (2016) stated that the power of deeper network architectures resides in their capabilities in representation learning. From an unraveled view, Veit, Wilber, and Belongie (2016) interpreted the deep ResNet as an ensemble of many relatively shallow networks, arguing that the effective depth of a 110-layer ResNet is less than 35. Based on this insight, Zagoruyko and Komodakis (2016) built a 16-layer wide ResNet outperforming the 1202-layer narrow ResNet, where wider means increased number of filters per convolution. However, deciding the depth and width of a handcrafted CNN architecture requires expert knowledge and trial-and-error.

Network architecture search (NAS) is a promising approach to automate the process of CNN architecture design. Early work in NAS built fully-connected neural networks based on low-level neurons and their connectivity (Schaffer, Whitley, and Eshelman 1992; Stanley and Miikkulainen 2002). Until recently, NAS was explicitly applied to CNNs (Baker et al. 2016). In general, NAS algorithms for CNNs can be divided into two categories, including evolutionary algorithms (EAs) and methods based on reinforcement learning (RL) algorithms (Xie and Yuille 2017; Cai et al. 2018; Zhong et al. 2018). These methods have been proven to be effective in designing network architectures with competitive results on standard benchmarks. However, most of the algorithms from existing literature require painstakingly designed solution space and dedicated operators to achieve high search efficiency.

In this paper, we propose a simple and effective framework for CNN architecture search with genetic programming (GP), which is called GP-CNAS. Originally, GP was a variation of EA and was specialized in evolving computer programs and function expressions (Koza 1994; 2010; Walker 2001). The standard GP encodes the solutions as trees with selected residual blocks as terminals (leaf nodes) and the block assembling procedure defined by primitive functions (non-leaf nodes). In this way, GP can express complex logics that are composed of these functions such as network architectures. For GP-CNAS, we adopt this tree encoding to explicitly express CNN architectures with variable depth and width. Moreover, the tree encoding also facilitates flexible modification on CNN architectures with unified operators.

To evaluate the GP-CNAS framework, we conduct experiments on the CIFAR-10 dataset. The experimental results show that with very limited CNN candidate validation, GPCNAS can find competitive CNN architectures compared to the state-of-the-art models.

The contributions of this paper are:

- The proposed GP-CNAS framework effectively encodes CNN architectures as trees, where subtrees in different sizes correspond to CNN substructures at different scale. The tree-based representation enables easy design and flexible implementation of genetic operators.
- During the genetic evolution, the dynamic crossover operator strikes a balance between exploration and exploitation, which emphasizes CNN complexity at early stage and CNN diversity at later stage. The desired CNN with balanced depth and width can be found within limited trials.
- The GP-CNAS framework belongs to the semi-automatic NAS algorithms, with CNN blocks as tree terminals. Our framework is highly extensible with other CNN blocks.

## Related Work

Designing network architectures remains to be a big challenge. Along with the development of handcrafted architectures, automatic network architecture search methods also underwent significant development over decades.

### EA for NAS

Early work in NAS employed EAs to optimize both the network architecture and the weights of the network (Miller, Todd, and Hegde 1989; Stanley and Miikkulainen 2002). Recently, gradient-based methods for weight optimization becomes a convention when EAs are solely used for optimizing network architectures (Lorenzo 2018). In the context of NAS for CNNs, EAs themselves mainly differ in network architecture representations and how they generate offspring. These representations can be divided into two categories: direct encoding like the directed acyclic graph (Real et al. 2017; Liu et al. 2018; Sun et al. 2018) and indirect encoding like the fixed-length binary string (Xie and Yuille 2017). The graph-style encoding simplifies the mapping from the solution space to representation space, while the string encoding can adopt genetic operators from existing EA literature. However, methods in the first category have difficulty in designing efficient genetic operators, and methods in the second category have difficulty in ensuring the feasibility of modified solutions. To summarize, both have the problem of inefficient sampling in search. Consequently, large population size and enough generations are required to find satisfactory CNN architectures.

### RL for NAS

Another family of methods exploits the reinforcement learning (RL) approach. Recent representative work include MetaQNN (Baker et al. 2016), EAS (Cai et al. 2018) and BlockQNN (Zhong et al. 2018). MetaQNN applied the Q-Learning paradigm with epsilon-greedy strategy to train a policy which sequentially chooses layer types and corresponding hyper-parameters. However, due to low sample efficiency, MetaQNN limits itself in designing small network architectures (Zhong et al. 2018). To overcome this problem, (Cai et al. 2018; Zhong et al. 2018) employed the macro structure of well-known handcrafted architectures and restricted the search space at the micro level (i.e. the inner-block architecture). Results indicate that RL-based methods can achieve competitive performance on standard datasets. To summarize, the design of small network architecture and inner-block architecture search is suitable for RL-based NAS.

### GP for NAS

GP is a variation of EA and specialized in evolving complex logics. (Koza and Rice 1991) successfully applied GP to find the optimal architecture of a neural network for the first time. They employed LISP symbolic expressions to represent various network architectures. Terminals in their GP framework adopted the finest network granularity, including the number of layers, the number of units per layer, and their connectivity. To our knowledge, the most related research to our work is by (Suganuma, Shirakawa, and Nagao 2017). Given a set of basic building blocks (i.e. ConvBlock and ResBlock), CNN structure and connectivity represented by Cartesian GP (CGP) encoding are optimized. Different from our work, CGP employed a directed acyclic graph with a two-dimensional grid to depict network structures. Due to the encoding complexity, only a mutation operator that randomly changes the type and connections of each block is applied. As shown in their experiment, the computational time to evolve small networks is relatively long.

## Method

In this section, we firstly introduce our tree-based encoding for CNN architectures, which is simple and extensible. Then, the overall framework of GP-CNAS is described in detail.

### GP Tree Representation of CNN Architectures

GP deals with the representation problem by adjusting the complexity of the structure undergoing adaptation (Koza 1994). A classical tree-based GP encoding is applied to denote CNN architectures in our method. The tree is composed of terminals (leaf nodes) and primitive functions (non-leaf nodes). Terminals consist of four residual blocks from (Zagoruyko and Komodakis 2016) based on their accuracies. Additionally, we define four operators as primitive functions that can reorganize CNN architectures in terms of depth and width.

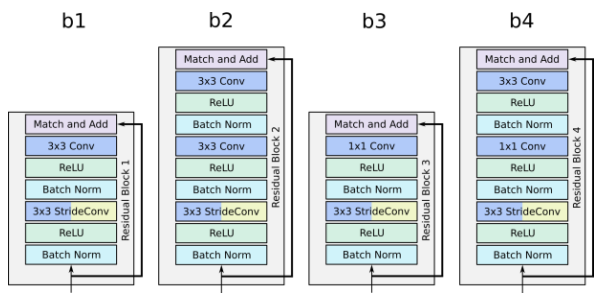

Figure 2: Four residual blocks selected from (Zagoruyko and Komodakis 2016)

**Terminals** Four residual blocks used in our CNN architecture search are illustrated in Figure 2, which are denoted as b1, b2, b3, b4. The shortcut connection not only accelerates training but also alleviates the vanishing gradient problem and the diminished feature reuse problem (Szegedy et al. 2017; Huang et al. 2017), which enables the training of deeper CNNs. According to previous research (He et al. 2016b), extra operations along the shortcut connection lead to higher training error. Therefore, no operation is added to the shortcut connection for our residual blocks.

In the body of each residual block, the Rectified Linear Unit (ReLU) (Nair and Hinton 2010) is selected to introduce non-linearity, which mitigates the vanishing gradient problem (Glorot, Bordes, and Bengio 2011) and has lower computational cost (Krizhevsky, Sutskever, and Hinton 2012). Batch Normalization desensitizes the ReLU after it to the scale variance of incoming signals (Ioffe and Szegedy 2015). Apart from that, the functionality of down-sampling is implemented in the convolutional layer with increased stride size, which eliminates the need for the pooling layer (Springenberg et al. 2014). It also can be noticed that only 3×3 and 1×1 filters are used for convolutions. The use of filters no larger than 3×3 is supported by (Szegedy et al. 2016), which stated that equivalent representational power can be achieved with lower dimensional embeddings in spatial aggregation. For the 1×1 filter, it is used for increasing or reducing the number of channels (Lin, Chen, and Yan 2013; Szegedy et al. 2015), which works as a fully-connected layer on the channel dimension.

**Primitive Functions** There are four types of primitive functions: (1) `^2` doubles the number of filters. We define multiple `^x` operations as $^{\sum_{x \in X} x}$, where X is the assemble of `^x` operators following the path from the root node to the target leaf node. As we set the basic filter number per convolution to 16, when `^2` acts on a block, the number of filters is $16 \times 2 = 32$; when three `^2` act on a block, the number of filters is $16 \times (2 + 2 + 2) = 96$. (2) In a similar way, `^3` triples the number of filters in target blocks. (3) + connects two blocks. (4) str doubles the size of stride for the target block. Both `^2` and `^3` make our CNN wider while + allow the architecture to become deeper. Additionally, the functionality of str is down-sampling.

For convenience and to guarantee the feasibility, the root node is set to + and the total number of str is restricted within 5. Moreover, only terminal nodes can be added after the primitive function str.

We can also express the tree as the *LISP symbolic expression*. Taking Figure 1 as an example, the tree can be expressed as (+(^3(^2 b2)(+b1(str b3)))). Correspondingly, this expression can also be translated into the network architecture as follows: (1) three blocks b2, b1, b3 are connected sequentially; (2) the number of filters of b2 can be calculated as $16 \times (2 + 3) = 80$ and others are default as 16; (3) the stride size of b3 is $1 \times 2 = 2$ and others are set as default 1.

### GP-CNAS Framework

In the evolutionary process, each CNN architecture is encoded as a tree by GP-CNAS where the validation accuracy after training is taken as its fitness. The evolutionary process of GP-CNAS can reproduce better CNNs generation by generation. The CNN architecture with the highest fitness is selected as the output of GP-CNAS. In this section, we describe the whole GP-CNAS framework at the beginning, which is shown in Algorithm 1. Then each part of GP-CNAS is detailed according to the logic flow of traditional evolutionary algorithms. Table 1 presents the specific types of operators that we implemented in GP-CNAS.

Table 1: GP-CNAS components

| Operator | Type | Reference |
|---|---|---|
| Initialization | Ramped-half-and-half | (Koza 1994) |
| Selection | Tournament | (Koza 1994; Koza and Poli 2005) |
| Crossover | Single-point crossover | (Langdon and Qureshi 1995) |
| Mutation | Subtree mutation | (Koza 1994; Koza and Poli 2005) |
| Update | Elitism | (Eshelman 1991) |

**Initialization** Firstly, a population with the pre-defined size is initialized with *ramped-half-and-half* (Koza 1994), which creates a population by two common GP tree-construction methods, which are *grow* and *full*. This mechanism generates an initial population of individuals with various sizes and structures. We define the *max_depth* as the maximum depth for all trees in the initial population. In ramped-half-and-half, the population is divided into *max_depth*−1 parts. The maximum depth of the trees constructed in each part is specified as {2, 3,... , *max_depth*} respectively. Each part contains half *grow* individuals and half *full* individuals. The *grow* process randomly selects a primitive function or a terminal as the current node from root to leaf sequentially. The construction process is terminated when we reach *max_depth* of trees or no new node can be appended because all leaf nodes are terminals. The *full* method builds the tree with elements from the function node set until reaching *max_depth* − 1. Then the process completes the tree by adding terminal nodes at the end.

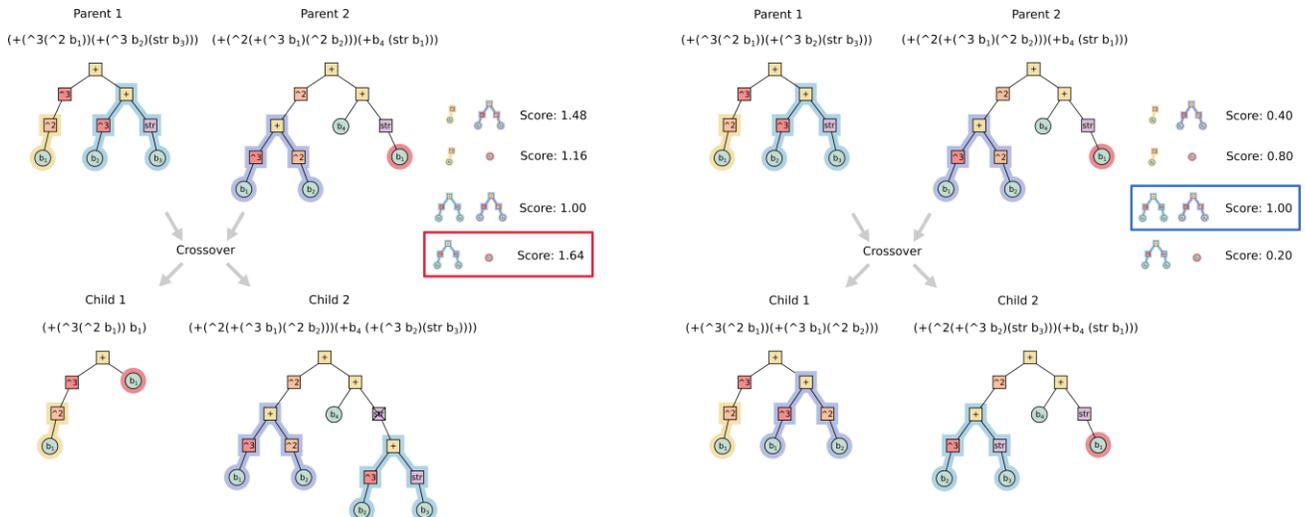

Figure 3: The emphasis of the crossover operator is different between the first $T/2$ generations and the last $T/2$ generations. For the first generation $t = 1$ as the left diagram shows, the subtree pair in the red box has the largest difference in node number (i.e. 4) hence has the highest score of 1.64. For the last generation $t = 10$ as the right diagram shows, the subtree pair in the blue box has the smallest difference in node number (i.e. 0) hence has the highest score 1.0. It can be noticed that the infeasible $str$ is deleted after crossover operation in child 2 of left diagram.

**Selection**  After population initialization, we perform the tournament selection (*line 6*) with a dynamic tournament size $\kappa$, defined as $\kappa = \lceil 2 + (N/2 - 2) \log_2(t)/\log_2(T) \rceil$, where $N$ denotes the population size, $T$ is the maximum number of generations and $t$ represents the current generation. The winners from two separate tournaments are selected as the parents for crossover. This function smoothly shifts the preference of tournament selection from the randomly selected individual to the best selected individual in the course of evolution.

**Crossover**  The task complexity affects the desired network size, that is the number of effective parameters. In order to aggressively approach desired network complexity for a given task and gradually transition to exploring network architectures, we design a dynamic crossover operator.

As shown in Algorithm 2, the crossover operator selects a pair of crossover points from parent trees according to the current generation $t$ and the difference in subtree sizes which is the number of nodes. A *roulette-wheel* method is applied to select a pair of subtrees according to the normalized scores (*line 9*). As Figure 3 shows, the score function guarantees that the subtree pair with large difference gets high score in the first $T/2$ generations and low score in the last $T/2$ generations. A subtree swap is executed after determining the crossover point in each tree.

**Mutation**  After crossover, we perform the mutation operation on all child individuals with the constant mutation rate $\mu$. In the mutation operation, a terminal node is randomly selected from the tree and replaced by a newly generated subtree using the *grow* method. The aim of mutation is to always explore larger networks with a small probability during the search process.

**Evaluation**  During the evolutionary process, the validation accuracies after training are taken as fitness of the CNN architecture.

**Elitism Update**  We apply an elitism mechanism to reserve the best $N$ individuals from $N$ parents and $N$ children. This mechanism prevents losing dominant individuals during the evolutionary process.

## Experimental Settings

### Data Preparation

All our experiments are conducted on the CIFAR-10 dataset (Krizhevsky and Hinton 2009). CIFAR-10 is a dataset for

**Algorithm 1:** GP-CNAS Framework

**Inputs**: maximum generation number $T$; population size $N$; mutation rate $\mu$
**Outputs**: $best\_individual$

1  $pop :=$ **initialize** $(\emptyset)$
2  $t := 1$
3  **while** $t \leq T$ **do**
4    $crossover\_pop := \emptyset$
5    **while** $size(crossover\_pop) \leq N$ **do**
6      $parents \leftarrow$ **select** $(pop)$
7      $children \leftarrow$ **crossover** $(parents, t)$
8      $children$ **add_to** $crossover\_pop$
9    $mutate\_pop := \emptyset$
10   **for** $child \in crossover\_pop$ **do**
11     **if** $rand\_num \leq \mu$ **then**
12       $child \leftarrow$ **mutate** $(child, t)$
13     $child$ **add_to** $mutate\_pop$
14   $accuracy \leftarrow$ **train_and_validate** $(mutate\_pop)$
15   $pop \leftarrow$ **update** $(mutate\_pop, pop)$
16   $t \leftarrow t + 1$
17  $best\_individual \leftarrow$ **select_best** $(pop)$

**Algorithm 2:** Crossover

**Definition:** ;
$node_i^k$, the $i^{th}$ node in the tree $k$;
$subtree_i^k$, subtree rooted at node $i$ in the tree $k$
**Inputs**: $parent\_tree_k, parent\_tree_m, t, T$
**Outputs**: $child\_tree_k, child\_tree_m$

1  $diff\_vec := \emptyset$
2  **for** all $node_i^k \in parent\_tree_k$ **do**
3    **for** all $node_j^m \in parent\_tree_m$ **do**
4      $value_{ij}^{km} \leftarrow diff(subtree_i^k, subtree_j^m)$
5      $value_{ij}^{km}$ **add_to** $diff\_vec$
6  $diff\_vec \leftarrow$ **normalize**$(diff\_vec)$
7  $score\_vec := \emptyset$
8  **for** all $value_{ij}^{km} \in diff\_vec$ **do**
9    $score_{ij}^{km} \leftarrow 1 + value_{ij}^{km} \cdot (T - 2t)/T$
10   $score_{ij}^{km}$ **add_to** $score\_vec$
11  $crossover\_point_{ij} \leftarrow$ **select** $(score\_vec)$
12  $child\_tree_k, child\_tree_m \leftarrow$ **swap_subtree** $(parent\_tree_k, parent\_tree_m, crossover\_point_{ij})$

image classification, which consists of 50,000 training samples and 10,000 test samples. Each sample is a 32×32 RGB image, and the task is to classify images into 10 categories. The original training set is randomly split into 40,000 training images and 10,000 validation images. During the network architecture search, all networks are evaluated on the validation set, and the validation accuracies are taken as their fitness. The test set is kept untouched during the network architecture search stage and only used once for the fittest individual. Conventional data augmentation strategies (Krizhevsky, Sutskever, and Hinton 2012), including horizontal flips, random crops from padded images (i.e. 2-pixel padding on each side with zeros) and sample-wise standardization, are also adopted for data preprocessing. Recently, new data augmentation methods, such as cutout (DeVries and Taylor 2017) and random erasing (Zhong et al. 2017) were developed and obtained improved accuracies on image recognition tasks. However, data augmentation of this type is beyond the scope of our discussion, hence not employed for fair comparison.

**Neural Network Training**

Apart from the network architecture, all hyper-parameters are identical throughout our experiments. All our networks are trained, evaluated and tested with mini-batch stochastic gradient descent with Nesterov momentum (Nesterov 1983), in which the momentum term is fixed at 0.9. The batch size is set to 128, with the training batch queried from a random shuffle queue, and both the validation batch and the test batch queried from first-in-first-out queues. To restrict the training time for each network, we train them for 200 epochs with the learning rate schedule from (Zagoruyko and Komodakis 2016). Specifically, the initial learning rate is set to 0.1, then dropped to 0.02 after 60 epochs, 0.004 after 120 epochs and 0.0008 after 160 epochs. Weight decay of 0.0005 (Krizhevsky, Sutskever, and Hinton 2012) is also used to alleviate overfitting. 200 CNNs in total are trained and evaluated in our experiment, that is 10 generations with the population size of 20. Our experiments are run on 20 NVIDIA Tesla P100 GPUs for 9 days in total.

**GP-CNAS Settings**

The parameters for the GP part are listed as follows. The population size $N$ is set to 20 and the number of generations $T$ is set to 10 considering the amount of computational resources and experimental time. To initialize the population, *max_depth* of both *grow* and *full* methods are set to 10. For the mutation operation, the mutation rate μ is set as 20%. In addition, the maximum depth of the newly generated subtree is 4.

## Result and Analysis

In this section, the performance of CNNs on the CIFAR-10 dataset is evaluated and discussed. The CNN with the best validation accuracy is selected and tested on the test set after training. The best architecture obtained from our GP-CNAS has 14 blocks, with 31 convolutional layers and 9.7 million parameters, and the final test accuracy is 94.57%.

**Performance Comparison** The performance of the GP-CNAS evolved network is compared with other state-of-the art network architectures in Table 2. These networks are categorized into three types, which are handcrafted networks, networks generated by semi-automatic algorithms and networks generated by fully-automatic ones, depending on how much human labor are involved in the network design. Since the building blocks are pre-defined while block parameters and network connections are learned, GP-CNAS can be categorized as a semi-automatic network architecture search algorithm. The final test accuracy is 94.57%. It can be noticed that our final performance on the CIFAR-10 dataset is competitive compared with other well-known models, with the test accuracy higher than the ResNet (He et al. 2016a) from the handcrafted type and the MetaQNN (Baker et al. 2016) from the fully-automatic type.

Table 2: Performance Comparison of CNN Architectures on the CIFAR-10 Dataset.

| Type | Model | Para | Accuracy |
|---|---|---|---|
| Handcrafted | VGG (Simonyan & Zisserman 2014) | 15.2M | 92.06% |
| | NIN (Lin et al. 2014) | - | 91.19% |
| | ResNet (He et al. 2016) | 1.7M | 93.39% |
| | Stochastic Depth (Huang et al. 2016) | 10.2M | 94.77% |
| | WRN (Zagoruyko & Komodakis 2016) | 11.0M | 95.19% |
| | FractalNet (Larsson et al., 2016) | 22.9M | 94.76% |
| | DenseNet (k=12) (Huang et al., 2017) | 7.0M | 95.90% |
| Fully-Auto | MetaQNN (Baker et al., 2016) | 11.2M | 93.08% |
| | NAS v3 (Zoph & Le 2016) | 7.1M | 95.53% |
| | NASNet-A (Zoph et al., 2017) | 3.3M | 96.59% |
| | CNN-GA (Sun et al., 2018) | 2.9M | 95.22% |
| Semi-Auto | CGP (Suganuma et al., 2017) | 1.7M | 94.02% |
| | Block-QNN-S (Zhong et al., 2018) | 6.1M | 95.62% |
| | GP-CNAS (Ours) | 2.0M | 92.64% |
| | GP-CNAS (Ours) | 5.4M | 94.10% |
| | GP-CNAS (Ours) | 9.7M | 94.57% |

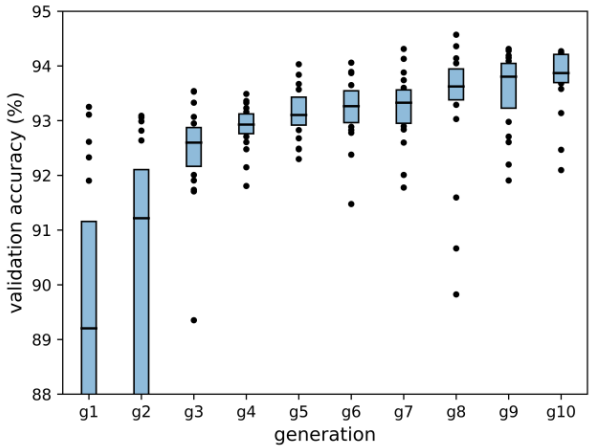

Figure 4: Boxplot of validation accuracies for all individuals of each generation. The upper and lower bounds of the box represent the first and third quartile of each generation, median value are shown as horizontal bars, and outliers are plotted as individual dots.

**Evolutionary Process** Regarding the evolutionary performance, the validation accuracies for all generations are shown in Figure 4. Specifically, the accuracy increases monotonically as the evolution proceeds. In the first generation, the *ramped-half-and-half* initialization scheme generates both shallow networks by the *grow* method and deep networks by the *full* method, which leads to the highest variance among all generations. Afterwards, those parents selected by the elitism mechanism are more likely to reproduce well-performed children. Moreover, the dynamic crossover operator strikes a balance between exploration and exploitation, which aggressively approaches the desired network complexity at early stage and proactively explore network diversity at later stage. Taking the advantages from the evolutionary process, CNN architectures are improved consistently and effectively.

The dynamic crossover operator and the elitism update mechanism bring the population towards the desired tree size. Figure 5 presents the number of tree nodes for all generations. In the first six generations, the averaged value increases steadily. While in the last four generations, the number of tree nodes approaches a value around 80 with decreased variance. It can be inferred that the model complexity reflected by the tree node number approaches the dataset complexity during the evolutionary process.

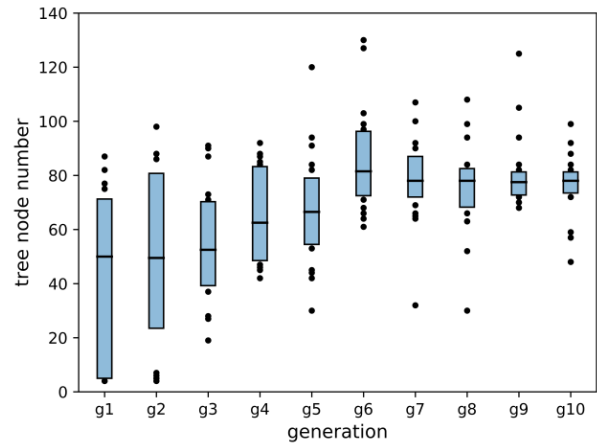

Figure 5: Boxplot of tree node number of all individuals for each generation. The upper and lower bounds of the box represent the first and third quartile of each generation, median value are shown as horizontal bars, and outliers are plotted as individual dots.

**Block Ratio** In Figure 6, the horizontal axis represents 200 individuals sorted by their fitness in ascending order. For each individual, its validation accuracy along with its block ratios is illustrated. It indicates that b2 and b3 are more likely to be the major block types in CNNs with high accuracies. For these "elite" networks on the right side of the graph, both block b2 and b3 have large ratio, with the averaged occurrence about 35%. In contrast, both block b1 and b4 are less likely to appear in the "elite" architectures, with the occurrence lower than 20%. Provided with the high extensibility of our GP-CNAS framework, the superior blocks can be detected given a set of candidate block types by the evolutionary process.

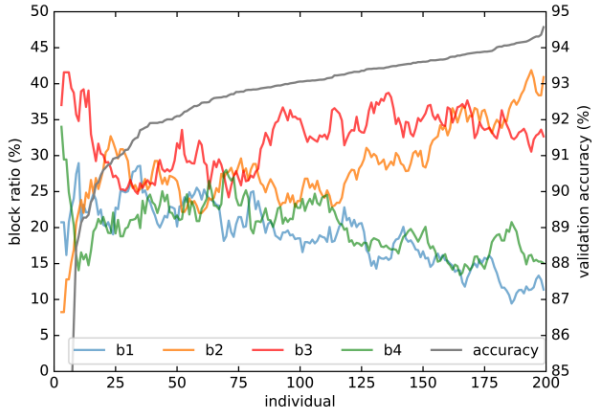

Figure 6: Block ratio and accuracy of each individual. The horizontal axis represents all individuals sorted by their fitness (i.e. gray line) in ascending order. The four colored lines depict the corresponding block ratio of each individual.

**Best Tree** The best tree obtained by GP-CNAS is shown in Figure 7. It contains 14 blocks, specifically b1×1, b2×7, b3×4 and b4×2. The total number of convolutional layers is 31 and the parameter size of the CNN is 9.7 million. In terms of network architecture, the search space in our GP-CNAS supports the search of both deep and wide CNN architectures. As depicted in the tree, our best CNN architecture shares some commonalities with WRN (Zagoruyko & Komodakis 2016). The network is relatively shallow with only 14 blocks. Meanwhile, some blocks contain a large number of filters. The widest block is the 11$^{th}$ one that contains 16×26 filters. Widening of blocks provides an effective way to reduce depth when the number of parameters is fixed, hence alleviates the gradient diminishing problem and facilitates efficient parallel computation. Similar to WRN, the number of strides is three in our best tree as well, with the stride locations evenly spaced in the latter half of the architecture.

## Conclusion and Future Work

In this paper, we propose the GP-CNAS framework to perform convolutional neural network architecture search with genetic programming. GP-CNAS encodes convolutional neural network architectures in trees, which enables effective and efficient search for optimal architectures with balanced depth and width. Based on the tree encoding, we design the dynamic crossover operator to aggressively approach desired network complexity and gradually transition to exploring network diversity. Our experimental results on the CIFAR-10 dataset confirmed our design intention.

Compared with encoding schemes such as the binary string and the directed acyclic graph, our tree representation enables easy design and flexible implementation of genetic operators. The GP-CNAS framework can automatically find competitive convolutional neural network architectures compared with other state-of-the-art models. Currently, the search space of GP-CNAS is restricted at the macro-level, where a set of handcrafted building blocks are provided in advance. In the future, the inner-block structure will be integrated into our tree representation, which leads to end-to-end NAS solutions.

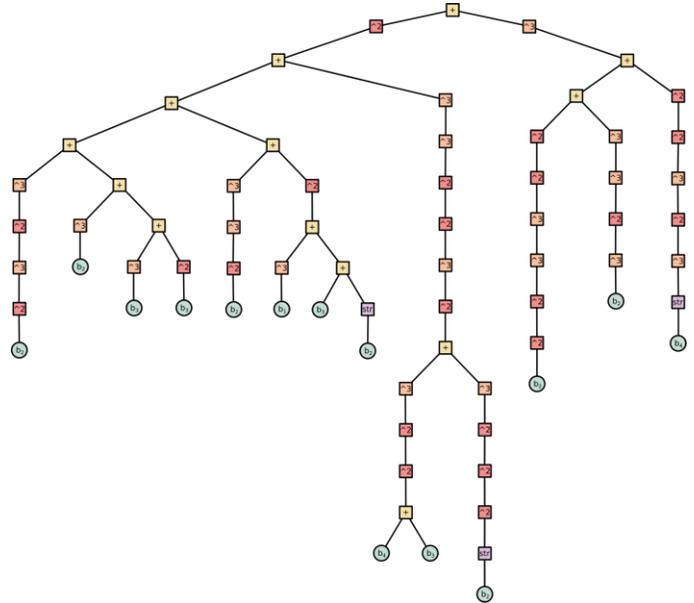

Figure 7: Architecture of the best tree output by GP-CNAS

## Acknowledgments

We would like to show our gratitude to Prof. Hong Jeff from School of Management and School of Data Science, Fudan University for his valuable advices about the presentation of this paper. This work has been supported by Alibaba Cloud for providing GPU computational resources and technical support.